# Using Bayesian Networks to Identify the Causal Effect of Speeding in Individual Vehicle/Pedestrian Collisions

Gary A. Davis
Department of Civil Engineering
University of Minnesota
Minneapolis, MN 55455

## Abstract

Estimating individual probabilities of causation generally requires prior knowledge of causal mechanisms. For traffic accidents such knowledge is often available and supports the discipline of accident reconstruction. In this paper structural knowledge is combined with Bayesian network methods to calculate the probability of necessity due to speeding for each of a set of vehicle/pedestrian collisions. Gibbs sampling is used to carry out the computations.

## 1 INTRODUCTION

Selecting a speed limit involves balancing the mobility of motorists against the external costs that mobility imposes on other road users. In setting speed limits, current traffic engineering practice gives preference to the 85th percentile of the vehicle speed distribution, and so for streets where the 85th percentile speed exceeds the current posted speed limit it can be argued that the posted speed limit is too low. Where this is the case, one measure of the external cost imposed by motorists' choice of speed is the number of vehicle/pedestrian accidents that would have been prevented had the prevailing speed limit been obeyed. This measure is especially important for local and residential streets, where pedestrians are naturally to be expected. To predict the accident reduction resulting from a safety countermeasure, traffic engineers usually apply an externally estimated accident reduction factor (RF) to an observed or predicted accident count. RFs are in turn generally estimated as

$$RF = (r_B - r_A)/r_B$$

where $r_B$ and $r_A$ respectively denote accident rates estimated before and after the application of the countermeasure. Unfortunately, external estimates from studies that are well-enough designed to support causal interpretations tend to be rare (Davis 2000), and for pedestrian accidents on local streets even questionable estimates are not available (TRB 1998). Interpreting the accident rates as probabilities, an RF estimated from a well-designed study can then be interpreted as what Pearl calls a "probability of necessity," computed assuming minimal knowledge of how accidents actually occur (2000, p. 292). By representing traffic accidents using structural models, an alternative estimate of the effect of speeding, specific to the accident history of a given area, can be had by computing the probability that speeding was necessary for the occurrence of each accident in that area.

To show this, we can begin with Rubin's potential response model (Holland 1986). Let j=1,..,N index a set of vehicle/pedestrian conflicts, which include actually occurring collisions as well as all instances where a driver had to brake in order to avoid hitting a pedestrian. For each conflict define the potential response variables

$h1_j =$    1, if pedestrian j was actually stuck,
         0, if pedestrian j was not struck.

$h2_j =$    1, if pedestrian j would have been struck under strict adherence to posted speed limits,
         0, if pedestrian j would not have been stuck under strict adherence.

Assuming that enforcement of the speed limit causes obedience to the limit, the change in accident frequency that would result if speed limits had been enforced is

$$\Delta = \sum_{\{j:h1_j=1\}} (h1_j - h2_j) + \sum_{\{j:h1_j=0\}} (h1_j - h2_j)$$

with the first sum on the right hand side giving accidents prevented by enforcement while the second sum gives accidents caused by enforcement. If drivers who were normally traveling at or below the speed limit would not have increased their speeds under strict enforcement, the mechanics of vehicle braking implies that the monotonicity condition $h1_j \geq h2_j$ is plausible, and the total accidents caused by enforcement is zero. If accident investigations have produced data relevant to the conditions of each accident, by coupling these data with prior knowledge about accident mechanisms it may be possible to compute individual probabilities of necessity, $P[h2_j=0 \mid h1_j=1, y_j]$, where $y_j$ denotes the data available for accident $j$. This leads to an estimate of the accident reduction

$$\hat{\Delta} = \sum_{\{j:h1_j=1\}} E[h1_j - h2_j \mid h1_j = 1, y_j] = \sum_{\{j:h1_j=1\}} P[h2_j = 0 \mid h1_j = 1, y_j]$$

Issues concerning individual accident causation frequently arise in legal contexts, where a jury must assess "cause in fact" (Robertson 1997), and so it is not surprising that over the past 60 years the discipline of traffic accident reconstruction has developed primarily to assist the legal system in resolving such issues. The definition of cause generally accepted in accident reconstruction work has been given by Baker (1975), who first defines a causal factor as "... any circumstance contributing to a result without which the result could not have occurred," (p. 274) and then defines the cause of an accident as the complete combination of factors which, if reproduced, would result in another identical accident (p. 284). This is arguably a deterministic view of accident occurrence, coupled with a counterfactual notion of causality. In accident reconstruction, physical models describing the behavior of entities involved in the accident are combined with information collected at the accident scene in order to infer the prior conditions leading to the accident. Causal assessment in accident reconstruction then often involves determining if, other things being equal, changes in some prior condition would have prevented the accident. Such assessments become more difficult when the variables appearing in an accident reconstruction are underdetermined by the available measurements and constraints. For example, the kinematic equation $s=v^2/2a$ can be used to estimate a vehicle's initial speed $v$ from a measured skidmark $s$, but only if one knows the vehicle's braking deceleration $a$. If the value for a variable is uncertain, a method for assessing the effect of this uncertainty on the final conclusion is needed, and this has often been accomplished by re-doing the reconstruction for a range of possible variable values. However, if one is willing to grant that uncertainty can be described using probability measures (Lindley 1987), more powerful methods are available. In particular, Brach (1994) has illustrated how statistical differentials can be used to compute the variance of a reconstruction estimate, while Wood and O'Riordain (1994) have presented what was in effect a Bayesian Monte Carlo approach for determining posterior distributions for an accident's initial conditions. Davis (1999) subsequently showed how Gibbs sampling could be used to compute a posterior distribution for the initial conditions of a vehicle/pedestrian accident, given measurements made at the accident scene. To date though no one has attempted to assess the uncertainty attached to causal or counterfactual claims.

Uncertainty assessment in accident reconstruction can be seen as a special case of a method for causal analysis that has been formalized by Pearl and his associates (Balke and Pearl 1994; Pearl 2000). To apply Pearl's method to the pedestrian accident problem an analyst would have to specify a probabilistic causal model consisting of (a) sets of exogenous and endogenous variables, (b) a set of structural equations expressing each endogenous variable as a function of exogenous and/or other endogenous variables, and (c) a probability distribution over the exogenous variables. A directed acyclic graph (DAG) can be used to summarize the qualitative dependency structure among the model variables, producing a Bayesian network. Pearl's Theorem 7.1.7 (2000, p. 206) then describes how the probability of a counterfactual claim, given evidence, can be evaluated. Balke and Pearl (1994) have also shown how technical difficulties arising from the need to describe or store a posterior distribution can be circumvented by applying updating methods to a DAG model that has been augmented to include nodes representing counterfactual outcomes.

## 2 STRUCTURAL MODEL

We will consider pedestrian/vehicle collisions consistent with the following scenario, illustrated in Figure 1. The driver of a vehicle traveling at a speed of $v$ notices an impending collision with a pedestrian when the front of the vehicle is a distance $x$ from the potential point of impact. After a perception/reaction time of $t_p$ the driver locks the brakes, and the vehicle decelerates at a constant rate $fg$, where $g$ denotes gravitational acceleration. After a transition time of $t_s$ the tires begin making skid marks and the vehicle comes to a stop, leaving a skidmark of length $s1$. Before stopping, the vehicle strikes the pedestrian at a speed of $vi$, the pedestrian is thrown into the air and comes to rest a distance of $d$ from the point of impact. The pedestrian is injured, and the severity of injury can be classed as slight, serious, or fatal. In addition, if the pedestrian was struck after the vehicle began skidding, it may be possible to measure a distance $s2$ running from the point of impact to the end of the skidmark. The basic inference problem is to characterize the posterior uncertainty in causal variables such as $v$, $vi$ and $x$ given some subset of the measurements $d, s1, s2$, and the pedestrian's injury severity. Figure 2 displays a DAG summarizing the conditional dependence structure of the collision model.

To complete the model it is necessary to specify deterministic or stochastic relations for the arrows appearing in Figure 2, and prior distributions for the background variables $x$, $v$, $t_p$, $t_s$ and $f$. From the above description it follows that the impact speed is given by

$$vi = v, \text{ if } x < vt_p$$
$$vi = 0, \text{ if } x > vt_p + v^2/(2fg)$$
$$vi = \sqrt{v^2 - 2fg(x - vt_p)}, otherwise$$

The theoretical length of the skidmark is computed as the difference between the total braking distance and the distance traversed during the transition phase,

$$theoretic\ skidmark = v^2/(2fg) - (vt_s - fgt_s^2/2)$$

Garrot and Guenther (1982), describing results from a series of controlled braking tests, reported that the standard deviation of measured skidmarks tended to increase as the initial speed increased, and that the average coefficient of variation for the difference between the measured and theoretical skid lengths was approximately equal to 0.11. In the reconstructions described below the measured skidmark, denoted by $s1$, was assumed to be a lognormal random variable with underlying normal mean equal to the natural log of the theoretical length, and underlying normal variance equal to 0.01, giving a coefficient of variation for the measurement error in the skidmarks of about 0.10. The measured second skid, $s2$, was also assumed to be lognormal with underlying normal mean equal to the natural log of the theoretical value and underlying variance equal to 0.01.

Next, measured throw distances and impact speeds from 55 crash tests between cars and adult pedestrian dummies, tabulated in (Eubanks and Hill 1998), were used to fit a linear model relating the natural log of the measured throw to the natural log of the measured speed. The residuals from this fit passed a test for being normally distributed, and the fitted model took the form

$$\log(d) = b_0 + b_1 \log(vi) + e$$

where $d$ = throw distance (meters), $vi$ = vehicle speed at impact (km/h), $e$ = normal random variable with mean equal to zero and estimated variance equal to 0.06. Least-squares estimates of the regression coefficients (together with the corresponding standard errors) were $b_0$=-3.43 (0.30) and $b_1$=1.61 (0.09).

To model the relationship between impact speed $vi$ and degree of injury severity an ordered logistic regression model was fit to data published in (Ashton 1980). The published data consisted of cross-tabulations of accidents, for three different pedestrian age groups, of vehicle impact speed versus pedestrian injury severity. The form of the fitted model was:

$$P[slight\ injury \mid vi] = L(b \cdot vi - a_1)$$
$$P[serious\ injury \mid vi] = L(b \cdot vi - a_2) - L(b \cdot vi - a_1)$$
$$P[fatal\ injury \mid vi] = 1 - L(b \cdot vi - a_2)$$

where $L(.)$ denotes the logit function. Weighted exogenous sampling maximum likelihood (WESML) was used to estimate model parameters, and a more detailed development of this model can be found in Davis (2001). The WESML estimates of the model parameters for adult pedestrians (ages 15-59), when the impact speed is in km/h, were $a_1$=4.07 (0.73), $a_2$=7.21 (1.01) and b=0.095 (0.02). Approximate standard errors are given in parentheses after each estimate.

Finally, the Bayesian network requires prior distributions for the background variables $x$, $v$, $t_p$, $t_s$ and $f$. Although prior distributions avoid the apparent arbitrariness of using fixed nominal values, they bring with them the problem of how to select these distributions in some reasonable manner. In deterministic sensitivity analyses it is often possible to identify defensible prior ranges for background variables (Niederer 1991), and Wood and O'Riordain argue that, in the absence of more specific information, uniform distributions restricted to these ranges offer a plausible extension of deterministic sensitivity methods (1994, p. 137). Following Wood and O'Riordian's suggestion, the reconstructions described in this paper used uniform prior distributions. Specifically, the range for $f$ was [0.45,1.0], and was taken from Fricke (1990, p. 62-14), where the lower bound corresponds to a dry, travel-polished asphalt pavement and the upper bound to a dry, new concrete pavement. The range for the perception/reaction time, $t_p$, was [0.5 seconds

, 2.5 seconds], which brackets the values obtained by Fambro et al. (1998) in surprise braking tests, and the midpoint of which (1.5 seconds) equals a commonly chosen nominal value (Stewart-Morris 1995). For the braking transient time, Neptune et al. (1995) reported values ranging between 0.1 and 0.35 seconds for a well-tuned braking system, while Reed and Keskin (1989) reported values in the range of 0.4-0.5 seconds, so the chosen range was [0.1 seconds, 0.5 seconds]. The strategy for the initial distance and initial speed was to use ranges wide enough so that no reasonable possibility was excluded a priori. The range for $v$ was [5 meters/second, 50 meters/second], and the range for $x$ was [0 meters, 200 meters]. Finally, uncertainty in the parameters for the throw distance and injury severity models was incorporated by treating the actual parameter values as normal random variables, with means and standard deviations equal to the estimates presented earlier. In Figure 2 the nodes labeled $p_1$ and $p_2$ represent these parameters.

## 3 APPLICATION

To estimate the accident reduction due to speed limit adherence one requires values for the individual probabilities of necessity, where $y_j$ denotes measurements of some subset of $s1$, $s2$, $d$ and the injury severity. These can be computed by (i) updating the distributions for $x$, $t_p$, $t_s$ and $f$ using the data $y_j$, (ii) setting the initial speed equal to the speed limit, and (iii) computing the probability that $vi=0$ (since $h2=0$ if and only if $vi=0$) using the updated distributions for $x$, $t_p$, $t_s$, and $f$ together with the condition that $v$ is set to the speed limit. For DAG models which have causal interpretations, Balke and Pearl (1994) describe how, by appropriately augmenting the DAG with additional nodes representing counterfactual variables, Bayesian network methods can be used to compute counterfactual probabilities. Figure 2 also shows the augmented DAG for the pedestrian collision model, where $v*$ and $vi*$ denote the counterfactual variables.

The application of this approach will be illustrated using data collected by Kloeden et al. (1997), where accident reconstruction methods were used to estimate the speeds of accident-involved vehicles as part of a study seeking to relate speed to accident risk. All the investigated accidents occurred on roads with a 60 km/h speed limit, but a sample of speeds from vehicles not involved in accidents showed an average speed of about 60 km/h, with an 85% percentile speed of about 80 km/h. Current traffic engineering practice would consider raising the speed limit to 80 km/h. The question at hand then is how many vehicle/pedestrian accidents would have been prevented had all vehicles obeyed the 60 km/h speed limit?

The sample of investigated accidents included eight vehicle/pedestrian collisions satisfying the conditions of the model described in Section 2, being frontal impacts by a single vehicle, with evidence of pre-impact braking. Information on each of the investigated accidents was published in volume 2 of Kloeden et al., from which measurements of $s1$, $s2$ and $d$, along the degree of injury suffered by the pedestrian, were obtained. Using the Gibbs sampling program BUGS (Gilks et al. 1994), posterior distributions were estimated, with the collision model described in Section 2 augmented to include the counterfactual variables $v*$ and $vi*$, and with $v*$ set equal to 60 km/h. A 5000 iteration burnin was followed by 50000 iterations with every 10th iteration being saved for analysis. Three separate Gibbs sampling chains were generated from different initial values and random number seeds, convergence was checked using the Gelman and Rubin test, and sample size was checked using the Raftery and Lewis test, as implemented in CODA (Best et al. 1995).

Figure 3 shows posterior means and 95% credible intervals for the vehicle initial speeds $v$, obtained from the Gibbs sampler, along with estimates made using two deterministic methods. Method 1 refers to estimates computed by applying the midpoints of the uniform priors to the braking model described in Section 2. Method 2 refers to the estimates given by Kloeden et al., which were computed using a somewhat different braking model and nominal input values. Overall, it appears that as long as one can find reasonable values for unmeasured input variables, the deterministic methods give reasonable approximations to the posterior mean speeds. It also appears that a reasonable prior uncertainty concerning input variables can induce a nontrivial degree of uncertainty in the resulting speed estimates.

Table 1. Gibbs Sampler and Deterministic Estimates.

|           | Gibbs Sampling |           | Deterministic $vi*$ |           |
|-----------|----------------|-----------|---------------------|-----------|
| Collision | P[v>60]        | P[vi*=0]  | Method 1            | Method 2  |
| 1         | 0.71           | 0.45      | 30                  | 30        |
| 2         | 0.90           | 0.76      | 0                   | 0         |
| 3         | 0.87           | 0.65      | 0                   | 8         |
| 4         | 0.64           | 0.43      | 26                  | 30        |
| 5         | 0.63           | 0.55      | 14                  | 14        |
| 6         | 0.42           | 0.29      | 21                  | 21        |
| 7         | 0.91           | 0.63      | 0                   | 11        |
| 8         | 0.09           | 0.03      | 29                  | 29        |

Table 1 shows the probability that each vehicle was initially exceeding the speed limit, along with assessments as to whether or not the vehicle would have stopped in time had it been travelling at the speed limit. Looking at the results from the Gibbs sampler, it appears that the vehicles in collisions 2, 3 and 7 were probably speeding and these accidents would probably have been prevented had the vehicle been traveling at 60 km/h. The vehicle in collision 8 was probably not speeding and the accident would probably not have been prevented by speed limit enforcement. For the remaining four cases the effect of speeding is uncertain. Looking at the two rightmost columns of Table 1 we see that deterministic method 1 predicts a reduction of three accidents due to speed limit adherence while method 2 predicts a reduction of only one accident. The $P[v_i^*=0]$ column of Table 1 gives the individual probabilities of necessity for each of the accidents, and summing the entries in this column gives an estimate of 3.8 accidents prevented by speed limit adherence. This can be interpreted as a measure of the accident reduction potential of speed limit enforcement.

## 4 CONCLUSION

Investigation and reconstruction of traffic accidents is often done to support criminal and civil legal proceedings and, to a lesser extent, to support traffic safety research. In all cases though, the ultimate objective is to use the evidence from an accident to identify or exclude possible causal factors, as rationally and objectively as is possible. Unlike more standard problems in causal inference, the underlying causal structure of a reconstruction problem can often be identified, but uncertainty arises because the initial conditions of the accident cannot be measured, and are usually underdetermined by the available evidence. To a greater or lesser extent then, the reconstructionist must supplement the evidence with prior knowledge concerning the values taken on by unmeasured variables, and uncertainties in this prior knowledge induce uncertainties in the estimates and conclusions produced by the reconstruction. At present there is no comprehensive or commonly accepted method for rationally accounting for this uncertainty, but three characteristics of this knowledge domain suggest that Bayesian network methods should be applicable. These are: (1) (local) Laplacean determinism, (2) a counterfactual view of causality, and (3) the use of probability measures to assess uncertainty. Results developed by Pearl and his associates can then be used to rigorously pose and answer selected counterfactual questions about an accident. Because reconstruction models often contain continuous variables and deterministic relationships, the exact updating methods developed for finite Bayesian networks are not at present well-suited to accident reconstruction, but approximations using Markov Chain Monte Carlo methods are more promising.

## Acknowledgements/Disclaimer

This research was supported by Minnesota's Local Road Research Board and the Minnesota Department of Transportation. However, all views expressed here are solely the responsibility of the author, and do not necessarily represent the positions of either of these agencies.

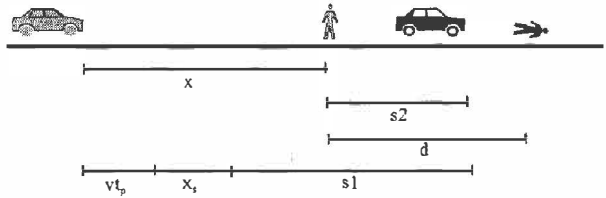

Figure 1. Diagram of a Vehicle/Pedestrian Collision. $x_s$ denotes the distance traveled by the vehicle during the transition phase.

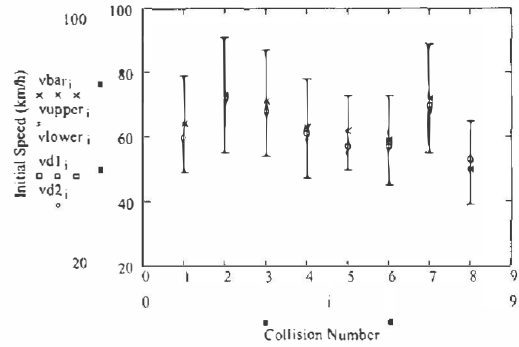

Figure 3. Posterior Means and 95% Credible Intervals for Initial Speeds in Eight Vehicle/Pedestrian Collisions. x's denote posterior means, squares denote deterministic method 1 estimates, diamonds denote deterministic method 2 estimates.

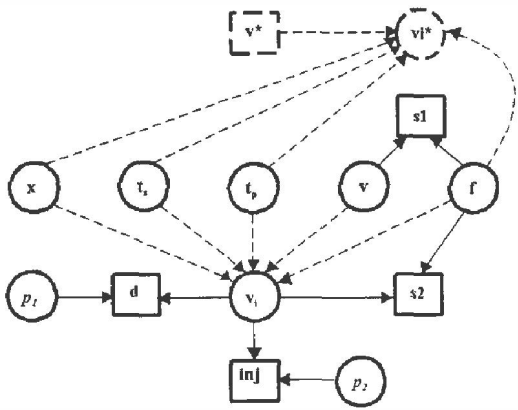

Figure 2. DAG Representation of a Vehicle/Pedestrian Collision. Circles denote unobserved variables, squares denote observed variables, dashed lines denote deterministic links, solid lines denote stochastic links.